\documentclass[12pt, letterpaper]{article}

\usepackage[utf8]{inputenc}
\usepackage{graphicx}
\usepackage{amsmath}
\usepackage{authblk}
\usepackage{amssymb}
\usepackage[margin=1in]{geometry}
\usepackage[utf8]{inputenc} 
\usepackage[T1]{fontenc}    
\usepackage{hyperref}       
\usepackage{url}            
\usepackage{booktabs}       
\usepackage{amsfonts}       
\usepackage{nicefrac}       
\usepackage{microtype}      
\usepackage{xcolor}         
\usepackage[
    backend=biber, 
    style=numeric, 
    sorting=none 
]{biblatex}

\title{A Foundational Multi-Modal Model for Few-Shot Learning}

\author[1]{Pengtao Dang}
\author[1]{Tingbo Guo}
\author[1]{Sha Cao}
\author[1]{Chi Zhang}

\affil[1]{Oregon Health \& Science University, Portland, OR, USA}

\date{}

\begin{document}

\maketitle

\begin{abstract}
  Few-shot learning (FSL) is a machine learning paradigm that aims to generalize models from a small number of labeled examples, typically fewer than 10 per class. FSL is particularly crucial in biomedical, environmental, materials, and mechanical sciences, where samples are limited and data collection is often prohibitively costly, time-consuming, or ethically constrained. In this study, we present an innovative approach to FSL by demonstrating that a Large Multi-Modal Model (LMMM), trained on a set of independent tasks spanning diverse domains, task types, and input modalities, can substantially improve the generalization of FSL models, outperforming models based on conventional meta-learning on tasks of the same type. To support this, we first constructed a Multi-Modal Model Few-shot Dataset (M3FD, over 10K+ few-shot samples), which includes 2D RGB images, 2D/3D medical scans, tabular and time-course datasets, from which we manually curated FSL tasks such as classification. We further introduced M3F (Multi-Modal Model for Few-shot learning framework), a novel Large Multi-Modal Model framework tailored for data-constrained scientific applications. M3F supports a wide range of scientific data types through a modular pipeline. By fine-tuning the model on M3FD, M3F improves model performance, making LMMM feasible for real-world FSL deployment. The source code is located at \url{https://github.com/ptdang1001/M3F} . To democratize access to complex FSL data and promote reproducibility for public usage, M3FD is paired with a flexible and user-friendly tool that enables efficient querying, task-specific sampling, and preprocessing. Together, our dataset and framework offer a unified, scalable solution that significantly lowers the barrier to applying LMMMs in data-scarce scientific domains.
\end{abstract}

\section{Introduction}
\label{introduction}

Many of the highest-value challenges in science and industry are characterized by profound data scarcity. While standard AI models achieve high performance by training on massive datasets, their application is severely limited in critical domains where data collection is expensive, slow, or ethically impossible\cite{bansal2022systematic, alzubaidi2023survey, dang2025physics, alghamdi2021graph}. These high-stakes fields—including healthcare for diagnosing rare diseases, industrial manufacturing for predicting unique equipment failures, and autonomous systems for handling rare edge-case scenarios—often have only 1 to 10 labeled examples available for a given class. This paradigm, known as Few-Shot Learning (FSL), aims to build reliable and generalizable models from such extremely limited data but has historically faced significant hurdles with overfitting and poor generalization\cite{guo2020broader,ge2023few}. While traditional solutions like meta-learning and data augmentation offer partial improvements, they often fall short in complex, real-world applications, especially when dealing with diverse data types\cite{wang2020generalizing, parnami2022learning, song2023comprehensive, sung2018learning}.

A new paradigm has emerged with the advent of Large Multi-Modal Models (LMMMs). Foundational works have demonstrated that models pre-trained on vast, web-scale datasets develop powerful emergent reasoning capabilities, enabling them to generalize to new tasks from very few examples, a concept powerfully illustrated by language models like GPT-4\cite{yin2024survey, liang2024survey, wei2022emergent}. Our vision is to leverage this power to create a foundational model that can learn from minimal data, unlocking solutions to problems previously deemed unsolvable due to data scarcity. We hypothesize that a single LMMM, trained on a diverse set of independent few-shot tasks, can achieve superior generalization compared to conventional methods.

Furthermore, the quality and diversity of the training data are paramount to this new paradigm\cite{yu2024makes, wettig2024qurating}. We operate on the principle that training a single, unified model on a wide spectrum of high-quality data—spanning multiple modalities (vision, text, tables), a variety of tasks, and diverse real-world scenarios—compels the model to learn a more abstract and robust underlying feature representation\cite{lee2023beyond, ramirez2022all, yu2023large, chung2023increasing, drosou2017diversity, dang2023generalized}. This creates a synergistic effect where knowledge gained from one data type can bolster the model's reasoning and generalization capabilities on another. This enhanced data efficiency and cross-domain knowledge transfer are critical for achieving state-of-the-art performance, particularly in the challenging context of few-shot learning.

This paper introduces a unified framework and a purpose-built dataset to address this challenge. Our contributions are threefold:
\begin{enumerate}
    \item We present the \textbf{Multi-Modal Model Few-shot Dataset (M3FD)}, a unique dataset of over 10k samples specifically curated for few-shot, multi-modal learning, covering a wide spectrum of data types including 2D/3D vision, tabular and time-course data.
    \item We propose the \textbf{Multi-Modal Model for Few-shot learning (M3F) framework}, a novel architecture built upon a state-of-the-art LMMM backbone that uses modality-specific encoders and a unified language interface to handle diverse data inputs.
    \item We introduce a novel \textbf{4-Stage Training Strategy}, which combines knowledge injection, curriculum learning with strategic input masking, and complex generation training to build robust, generalizable representations from sparse data.
\end{enumerate}

\begin{figure}[htp]
    \centering
    \includegraphics[width=0.8\linewidth]{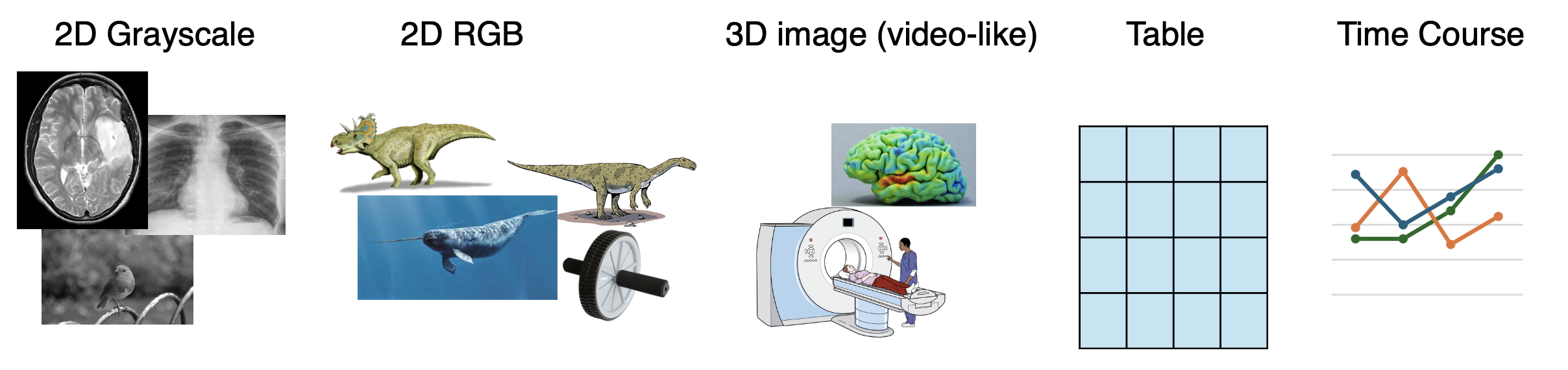}
    \caption{Data Example}
    \label{fig:data-example}
\end{figure}

\section{Related Work}
\label{related_work}

Our work is situated at the intersection of several key research areas: Few-Shot Learning, Large Multi-Modal Models, and Self-Supervised Representation Learning.

\paragraph{Few-Shot Learning (FSL).} Traditional FSL approaches often rely on meta-learning, or "learning to learn" Methods like Matching Networks \cite{vinyals2016matching} and Prototypical Networks\cite{ji2020improved, snell2017prototypical} learn a metric space where classification can be performed by computing distances to prototype representations of each class.While effective on benchmark tasks, these meta-learning strategies often fall short in complex, real-world scenarios, facing issues with robustness and generalization to out-of-domain data\cite{song2023comprehensive}. Transfer learning, another popular approach, fine-tunes a model pre-trained on a large source dataset on the small target dataset \cite{pan2010survey}. Our work extends this concept by using a massively pre-trained LMMM as the source model.

\paragraph{Large Models as Few-Shot Learners.} The paradigm shifted with the introduction of GPT-3 \cite{brown2020language}, which demonstrated that language models trained at an unprecedented scale develop "in-context learning" abilities, allowing them to perform new tasks given only a few examples in a prompt. This proved that broad prior knowledge is essential for few-shot performance\cite{khattak2023maple, zhou2022conditional}. This principle was extended to the visual domain with models like Flamingo \cite{alayrac2022flamingo}, which aligned pre-trained vision and language encoders to enable remarkable few-shot visual understanding. Our framework builds directly on this assumption by using a state-of-the-art LMMM, LLaVA-NeXT-Video \cite{liu2024llava-next-video}, as its foundation.

\paragraph{Self-Supervised Learning and Masked Autoencoders.} A key challenge in FSL is learning rich features from limited data\cite{song2023comprehensive}. Self-supervised learning methods provide a solution by creating pretext tasks directly from the input data\cite{hendrycks2019using}. A particularly effective strategy is masked input modeling. Masked Autoencoders (MAE) \cite{he2022masked} demonstrated that masking large portions of an image and training a model to reconstruct the missing pixels leads to superior, holistic feature learning. We adopt this principle of strategic input masking in our training strategy to force our encoders to learn generalizable features from sparse signals.

\section{The Multi-Modal Model Few-shot Dataset}
To effectively train and evaluate a model for multi-modal few-shot learning, we first constructed the Multi-Modal Model Few-shot Dataset (M3FD). Standard datasets are often large-scale for a single modality or lack the specific structure needed to test few-shot generalization across diverse data types. 

The M3FD, Fig \ref{fig:data-example}, is a high-quality dataset of over 10k samples, intentionally designed to address this gap. It features comprehensive modality coverage, including 2D grayscale images, 2D RGB images, 3D video-like volumetric scans (MRI, CT), tabular, and time-course data. The core principle of the dataset is its structure: the data is organized into many distinct classes, each containing only 1-10 labeled samples. This explicitly creates the challenging few-shot scenario required for our research.

Following our "Strength in Diversity" principle\cite{lee2023beyond}, every sample was manually reviewed and labeled to ensure a rich mixture of sources, scenarios, and characteristics, which is critical for model generalization and reducing bias. While every sample has a class label, a subset includes detailed descriptions to support complex generation tasks. To further enhance data diversity and quality at the training level, we developed several custom prompt templates. These patterns dynamically embed the class labels or detailed descriptions into varied instructional formats, which prevents the model from overfitting to a single prompt structure and improves its ability to follow instructions robustly. To promote reproducibility, the M3FD is paired with a user-friendly tool for efficient data querying, task-specific sampling, and preprocessing.

\begin{figure}[htp]
    \centering
    \includegraphics[width=0.8\linewidth]{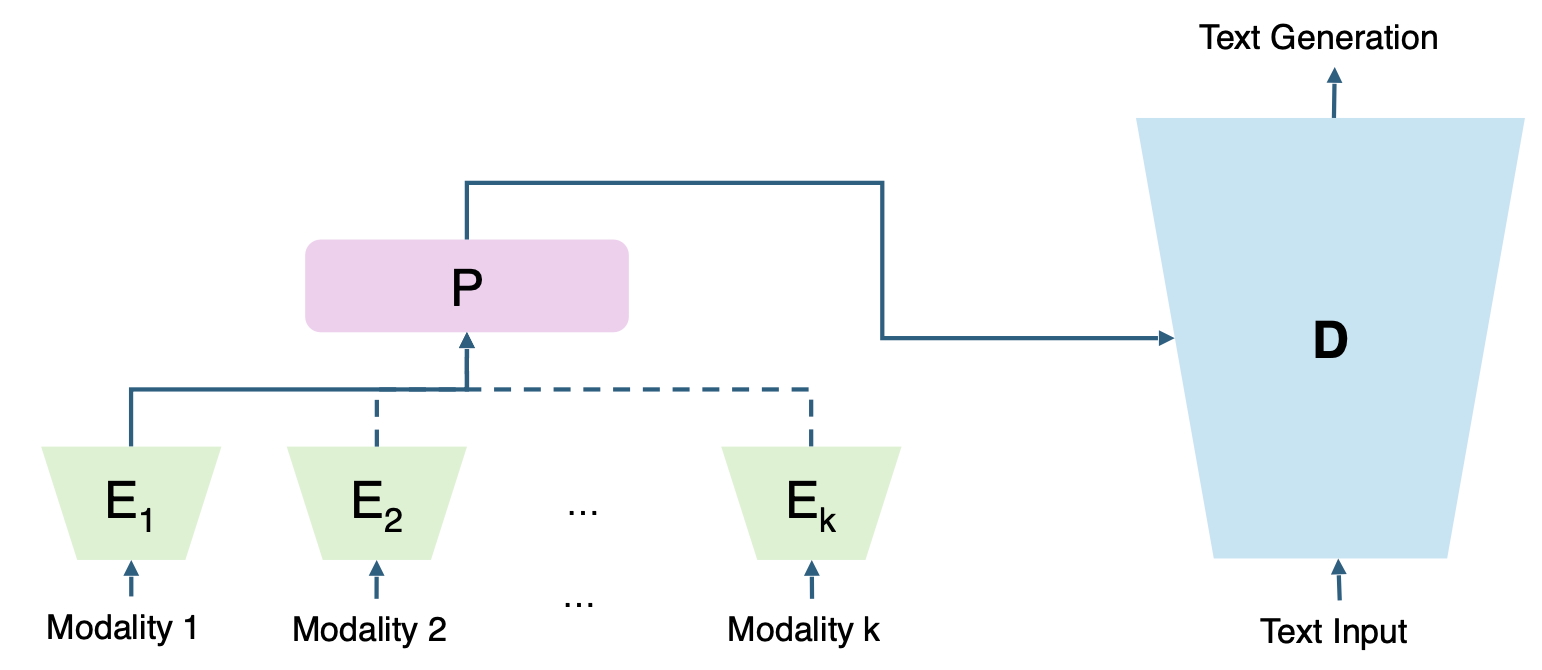}
    \caption{Multi-Modal Model for Few-shot learning Framework,}
    \label{fig:framework}
\end{figure}

\section{Methodology: The M3F Framework and Training Strategy}
We propose the Multi-Modal Model for Few-shot learning (M3F) framework, which leverages a unified architecture and a carefully designed 4-stage training process.

\subsection{Framework Architecture}
The M3F framework is designed to be modular and scalable, using text as a universal interface for all tasks and modalities. As illustrated in Fig \ref{fig:framework}, it consists of three main components:
\begin{itemize}
    \item \textbf{Modality-Specific Encoders:} Each input modality (e.g., 2D image, 3D volume, table, time-course) is processed by its own specialized encoder (E) designed to extract relevant features.
    \item \textbf{Multi-Modal Projector:} A projector network (P) maps the feature embeddings from all encoders into a common representation space that is compatible with the language model's vocabulary.
    \item \textbf{Language Decoder:} We use the decoder (D) of a powerful pre-trained LMMM as the backbone. For this work, we selected LLaVA-NeXT-Video for its state-of-the-art video and language reasoning capabilities. This allows us to inherit a vast base of knowledge and reasoning ability. All tasks, from classification to prediction, are framed as language generation problems, providing a single, flexible output interface.
\end{itemize}

\begin{figure}[htp]
    \centering
    \includegraphics[width=0.8\linewidth]{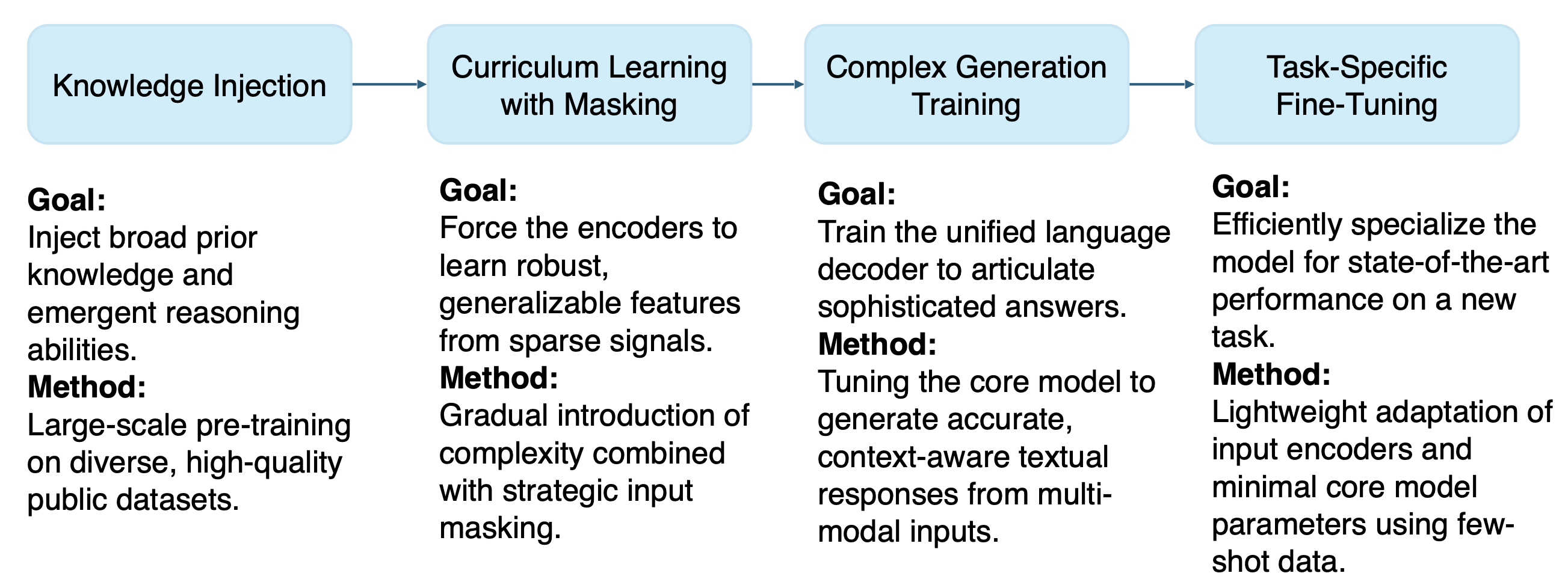}
    \caption{4-Stage Training Strategy}
    \label{fig:4-stage-training}
\end{figure}

\subsection{4-Stage Training Strategy}
As shown in Fig \ref{fig:4-stage-training}. We build the model's intelligence through a 4-stage process that moves from broad knowledge injection to task-specific specialization. 

\paragraph{Stage 1: Knowledge Injection.} The initial stage of our strategy is focused on Knowledge Injection, with the primary objective of instilling broad, multi-modal prior knowledge and emergent reasoning abilities into the foundational model. To achieve this, we continue training the full model on our M3FD dataset by framing the learning objective as a classification task. We selected this approach because classification provides a structured and highly efficient learning signal that is well-suited to the sparse nature of our data. This method simplifies the data preparation and preprocessing requirements, significantly lowering the barrier to incorporating our diverse set of few-shot modalities. The learning is guided by custom prompt templates that dynamically embed the new knowledge and class labels for each sample.

\paragraph{Stage 2: Curriculum Learning with Masking.} Following the initial knowledge injection, Stage 2 aims to force the modality-specific encoders to learn robust and generalizable features from sparse data signals. This is achieved through a novel combination of curriculum learning and a discriminative masking strategy applied across modalities. Specifically, we strategically hide parts of the input data based on its type: for images, we mask random patches; for tabular data, we simultaneously mask a subset of both rows and columns; and for time-course data, we mask specific time points and randomly selected features. To compel the model to learn a robust mapping from these partial signals, we introduce a special learnable embedding for each modality. This embedding is processed differently depending on the data type. For visual data, both the special embedding and the embeddings of the remaining visible patches are passed to the language decoder. For tabular data, however, we found it more effective to pass only the special embedding to the decoder, forcing it to act as a complete, holistic summary of the masked table. In all cases, this embedding is used in conjunction with the language decoder's output to predict the label of the masked input, forcing the model to learn a generalized representation that is not dependent on the specific distribution of any single data batch. This multi-faceted masking strategy is paired with a curriculum learning\cite{bengio2009curriculum} approach where data complexity is gradually increased.

\paragraph{Stage 3: Complex Generation Training.} The objective of Stage 3 is to transition the model's capabilities from discriminative classification to complex generative reasoning. This stage focuses on training the unified language decoder to articulate sophisticated and context-aware answers based on multi-modal inputs. For this, we leverage a curated subset of approximately 1k samples from our M3FD that are accompanied by detailed, long-form descriptions. Using complex prompts designed to elicit nuanced responses, we fine-tune the model to generate accurate and contextually appropriate text. To ensure this training efficiently hones the model's generative abilities without disrupting the robust features learned in the encoders during prior stages.

\paragraph{Stage 4: Task-Specific Fine-Tuning.} The final stage of our framework is the Task-Specific Fine-Tuning, designed to efficiently specialize the foundational model for state-of-the-art performance on a new target task using very few examples. To accomplish this while preserving the rich, generalized knowledge acquired in the prior stages, we "freeze" the vast majority of the core model's parameters. Adaptation is achieved through lightweight methods, primarily focusing on fine-tuning the relevant input encoders for the new task's data and a small number of core model parameters. This efficient approach allows the model to leverage its powerful pre-trained foundation to achieve high performance on a novel problem with minimal data and computational overhead.

\begin{figure}[htp]
    \centering
    \includegraphics[width=0.9\linewidth]{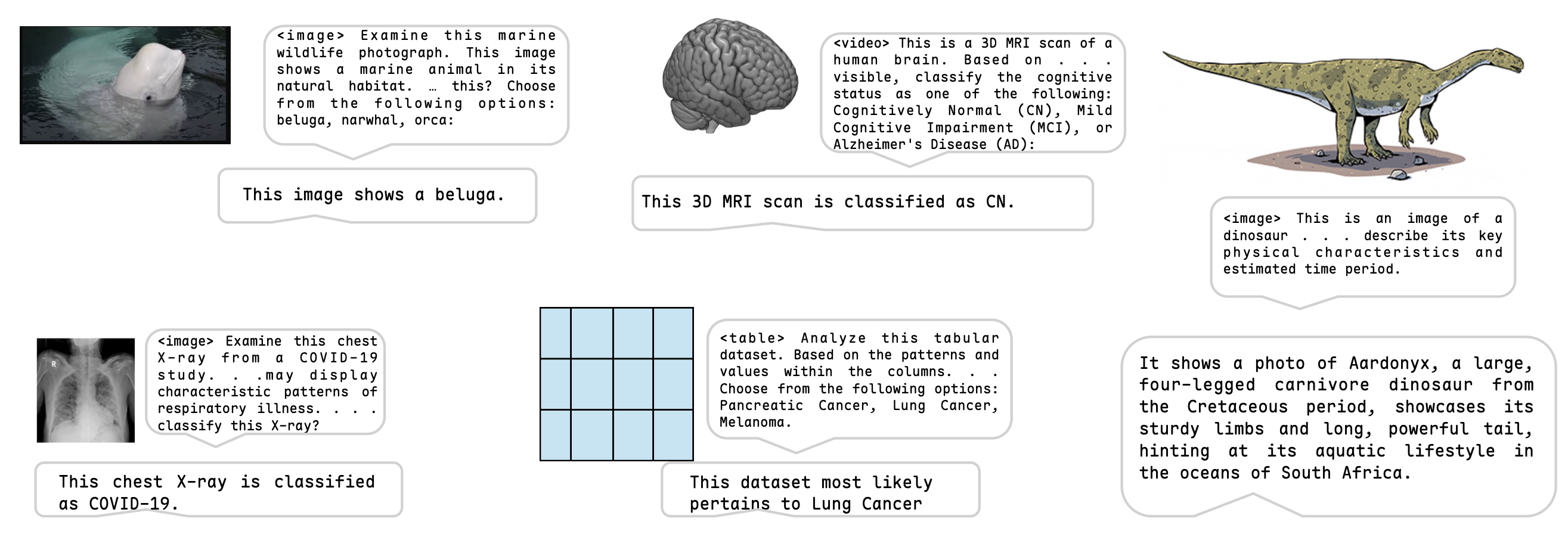}
    \caption{Output Example}
    \label{fig:output_example}
\end{figure}

\section{Experiments}
To validate the effectiveness of our M3F framework and the 4-stage training strategy, we conducted a series of experiments on our purpose-built M3FD dataset. The evaluation was designed to test the model's core few-shot learning capabilities across a diverse set of modalities and tasks.

\subsection{Experimental Setup}

\paragraph{Dataset.}
All experiments were performed on our \textbf{M3FD}. The dataset's structure is specifically designed for this challenge. This setup allows for a rigorous evaluation of the model's performance in extreme data-scarce conditions across its supported modalities: 2D grayscale/RGB images, 3D video-like scans, tabular, and time-course data.

\paragraph{Tasks and Evaluation.}
The evaluation spanned a range of few-shot tasks derived from the M3FD. This included multi-class classification (e.g., identifying wildlife, classifying chest X-rays) and simple generative tasks (e.g., describing an object's characteristics), as shown in our output examples Fig \ref{fig:output_example}. Performance was measured using task-appropriate metrics micro F-1 score for classification tasks.

\subsection{Baseline for Comparison}
To contextualize the performance of our M3F framework, we compared it against a classic and highly influential baseline from the few-shot learning literature: the Prototypical Network. This method was chosen as it represents a canonical meta-learning, or "learning to learn" approach, which is a traditional solution for few-shot tasks. Prototypical Networks operate by learning an embedding space where a "prototype" for each class can be computed by averaging the embeddings of the few available support samples. Classification is then performed for a query sample by finding the closest class prototype in this metric space. By benchmarking against this method, we directly compare our large model-based paradigm against the established meta-learning paradigm, evaluating their respective abilities to generalize from the sparse data in our M3FD dataset.

\subsection{Training Details}
All experiments were conducted on a single NVIDIA A100 80G GPU, utilizing bf16 mixed-precision. Our 4-stage strategy employed different fine-tuning techniques appropriate for each phase:
\begin{itemize}
    \item \textbf{Stage 1:} For the initial stage, we performed full-model fine-tuning to instill broad, foundational knowledge across the entire network.
    \item \textbf{Stages 2, 3, and 4:} For all subsequent stages—Curriculum Learning with Masking, Complex Generation Training, and Task-Specific Fine-Tuning—we employed a parameter-efficient approach using LoRA. Based on our ablation studies, a LoRA rank of 16 was used for these stages, as it provided the optimal balance between performance and training efficiency.
\end{itemize}

\subsection{Performance and Ablation Studies}
Our experiments confirm that our 4-stage training and masking strategy is highly effective, outperforming direct fine-tuning and the selected baselines.

\begin{figure}[htp]
    \centering
    \includegraphics[width=0.9\linewidth]{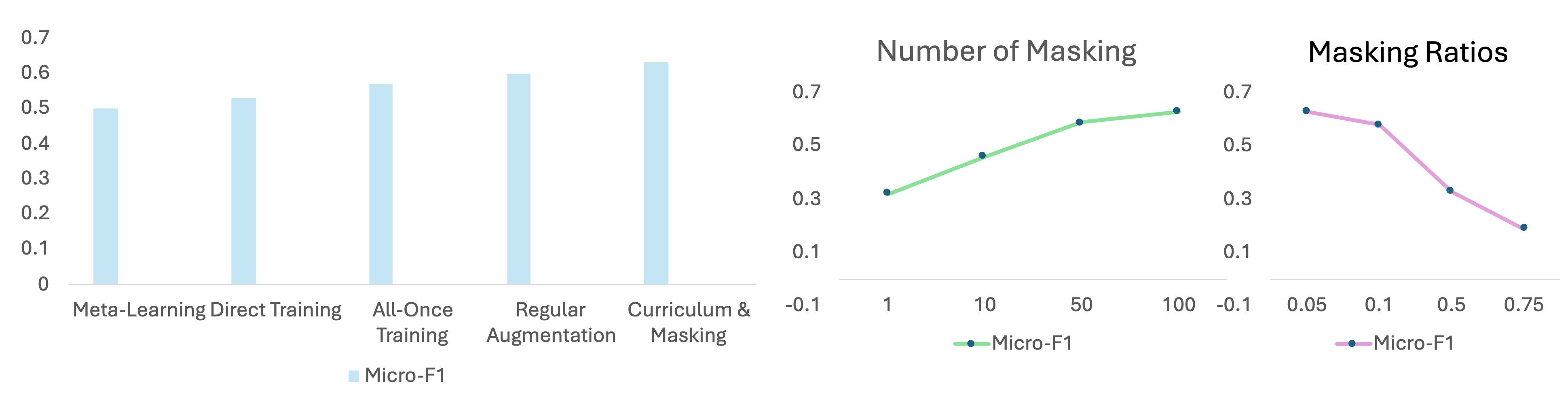}
    \caption{Experiment Results}
    \label{fig:experiment_results}
\end{figure}

\paragraph{Overall Performance.}
To evaluate our framework's effectiveness, we benchmarked our proposed Curriculum and Masking strategy against several alternative training approaches on the M3FD dataset. As shown in the experimental results Fig \ref{fig:experiment_results}, ours achieved a Micro-F1 score of approximately 0.63, delivering superior few-shot learning performance. This result demonstrates a clear improvement over traditional FSL meta-learning (Micro-F1 $\approx 0.50$), direct fine-tuning $(\approx 0.53)$, and standard data augmentation techniques $(\approx 0.60)$. This confirms that the combination of curriculum learning with our discriminative masking strategy is highly effective for building generalizable knowledge from sparse data.

\paragraph{Ablation Studies.}
We conducted comprehensive ablation studies to validate our key design choices.
\begin{itemize}

\item \textbf{Masking Strategy:} We analyzed the impact of our masking strategy during Stage 2 of training.
\begin{itemize}
\item \textbf{Number of Masking Applications:} We found that performance is directly correlated with the extent of exposure to the masking task. The Micro-F1 score steadily improved as the number of masking applications increased from 1 to 100, indicating that repeated training with partial signals is critical for learning robust features.
\item \textbf{Masking Ratio:} We investigated the optimal ratio of input to mask. Our results show that a relatively low masking ratio of 0.05 yielded the best performance (Micro-F1 $\approx 0.65$). Performance decreased as the ratio increased, suggesting that for our discriminative task, a small amount of masking provides a sufficient regularization signal to encourage contextual learning, while excessive masking removes too much critical information.
\end{itemize}

\end{itemize}

\section{Conclusion}
In this work, we developed a powerful and robust large multi-modal model designed specifically to excel at few-shot learning in data-scarce domains. Our framework is uniquely capable of processing a wide spectrum of data inputs. We demonstrated that our innovative 4-stage training strategy, which features curriculum learning and strategic input masking, is highly effective at building generalizable contextual knowledge from limited samples. The model's state-of-the-art performance in a challenging Alzheimer's prediction case study proves its adaptability and effectiveness on real-world problems. Despite these strong results, the model has limitations, primarily concerning the generation of complex, factually-dense descriptions. This is a direct consequence of the intentional sparsity of detailed descriptive data in our training set. For instance, in Fig \ref{fig:output_example}, when prompted to describe an image of a dinosaur, the model correctly identified the species as "Aardonyx" and accurately described its key physical characteristics like being a "large, four-legged" animal with "sturdy limbs and long, powerful tail". However, it incorrectly stated its living time as the "Cretaceous period" and hallucinated an "aquatic lifestyle in the oceans of South Africa". This highlights a key challenge: while our training strategy effectively teaches the model to recognize and describe visual patterns from few examples, instilling deep, factual knowledge for every niche domain remains a challenge constrained by the limited training data.


\printbibliography


\end{document}